\documentclass[sigconf]{acmart}

\usepackage[linesnumbered,ruled,vlined]{algorithm2e}
\usepackage{graphicx}
\usepackage{subcaption}
\usepackage{mwe}
\usepackage{color}
\usepackage{footmisc}
\usepackage{url}
\usepackage{tikz}
\usepackage{bm}
\usepackage{booktabs}
\usepackage[mathscr]{eucal}
\usetikzlibrary{arrows,positioning,automata,calc,shapes}
\usepackage{pgfplots}
\pgfplotsset{compat=newest, scaled z ticks=false} 
\pgfplotsset{plot coordinates/math parser=false}
\newlength\figureheight 
 \newlength\figurewidth
\newcommand{\red}{\color{red}}
\newcommand{\ra}[1]{\renewcommand{\arraystretch}{#1}}
\usepackage{enumitem}

\newcommand{\coin}{COIN}

\allowdisplaybreaks
\newcommand\numberthis{\addtocounter{equation}{1}\tag{\theequation}}

\begin{document}

\setcopyright{none}





%

\title{Contextual Outlier Interpretation}

%
%
%
%
%


\author{Ninghao Liu}
\affiliation{%
  \institution{Texas A\&M University}
  \city{College Station} 
  \state{Texas} 
}
\email{nhliu43@tamu.edu}

\author{Donghwa Shin}
\affiliation{%
  \institution{Texas A\&M University}
  \city{College Station} 
  \state{Texas} 
}
\email{donghwa_shin@tamu.edu}

\author{Xia Hu}
\affiliation{%
  \institution{Texas A\&M University}
  \city{College Station} 
  \state{Texas} 
}
\email{xiahu@tamu.edu}
      

\begin{abstract}
Outlier detection plays an essential role in many data-driven applications to identify isolated instances that are different from the majority. While many statistical learning and data mining techniques have been used for developing more effective outlier detection algorithms, the interpretation of detected outliers does not receive much attention. Interpretation is becoming increasingly important to help people trust and evaluate the developed models through providing intrinsic reasons why the certain outliers are chosen. It is difficult, if not impossible, to simply apply feature selection for explaining outliers due to the distinct characteristics of various detection models, complicated structures of data in certain applications, and imbalanced distribution of outliers and normal instances. In addition, the role of contrastive contexts where outliers locate, as well as the relation between outliers and contexts, are usually overlooked in interpretation. To tackle the issues above, in this paper, we propose a novel Contextual Outlier INterpretation (\coin) method to explain the abnormality of existing outliers spotted by detectors. The interpretability for an outlier is achieved from three aspects: outlierness score, attributes that contribute to the abnormality, and contextual description of its neighborhoods. Experimental results on various types of datasets demonstrate the flexibility and effectiveness of the proposed framework compared with existing interpretation approaches.




\end{abstract}

%
%
%

%
%


\maketitle
\section{Introduction}
Outlier detection has become a fundamental task in many data-driven applications. Outliers refer to isolated instances that do not conform to expected normal patterns in a dataset~\cite{Chan-etal09anomaly,Camp-etal16ontheevaluation}. Typical examples include notable human behaviors in static environment~\cite{Xiong-etal11direct}, online spam detection~\cite{liu2017detecting,shah2017flock,zhao2015enquiring}, public disease outbreaks~\cite{Wong-etal02rule}, and dramatic changes in temporal signals~\cite{Yang-etal02topic, Ma-Perk03online}. In addition, outlier detection also plays an essential role in detecting malevolence and contamination towards a secure and trustworthy cyberspace, including detecting spammers in social media~\cite{Yang-etal11free,Akog-etal10oddball} and fraudsters in financial systems~\cite{Phua-etal10comprehensive}.

Complementing existing work, enabling interpretability could benefit outlier detection and analysis in several aspects. First, interpretation helps bridging the gap between detecting outliers and identifying domain-specific anomalies. Outlier detection can output data instances with rare and noteworthy patterns, but in many applications we still rely on domain experts to manually select domain-specific anomalies out of the outliers that they actually care about in the current application. For example, in e-commerce website monitoring, outlier detection can be applied for discovering users or merchants with rare behaviors, but administrators need to check the results to select those involved in malicious activities such as fraud. Interpretation for the detected outliers, which provides reasons for outlierness, can significantly save the effort of such manual inspection. Second, interpretation can be used in the evaluation process to complement current metrics such as the area under ROC curve (AUC) and nDCG~\cite{davis2006relationship} which provide limited information about characteristics of the detected outliers. Third, a detection method that works well in one dataset or application is not guaranteed to have good performance in others. Unlike supervised learning methods, outlier detection is usually performed using unsupervised methods and cannot be evaluated in the same way. Thus, outlier interpretation would facilitate the usability of outlier detection techniques in real-world applications.

To this end, one straightforward way for outlier interpretation is to apply feature selection to identify a subset of features that distinguish outliers from normal instances~\cite{Knor-Ng99findingintensional,Mice-etal13explaining,Duan-etal14mining,Vinh-etal16discovering}. However, first it is difficult for some existing methods to efficiently handle datasets of large size or high dimensions~\cite{Vinh-etal16discovering}, or effectively obtain interpretations from complex data types and distributions~\cite{Knor-Ng99findingintensional}. Second, we measure the abnormality degrees of outliers through interpretation, which is important in many applications where some actions may be taken to outliers with higher priority. Some detectors only output binary labels indicating whether each data instance is an outlier. Sometimes continuous outlier scores are provided, but they are usually in different scales for different detection methods. A unified scoring mechanism by interpretation will facilitate the comparisons among various detectors. Third, besides identifying the notable attributes of outliers, we also analyze the context (e.g., contrastive neighborhood) in which outliers are detected. ``It takes two to tango." Discovering the relations between an outlier and its context would provide richer information before taking actions to deal with the detected outliers in real applications.


To tackle the aforementioned challenges, in this paper, we propose a novel Contextual Outlier INterpretation (\coin) approach to provide explanations for outliers identified by detectors. We define the interpretation of an outlier as the triple of noteworthy features, the degree of outlierness and the contrastive context with respect to the outlier query. The  first two components are extracted from the relations between the outlier and its context. Also, the interpretations of all outliers can be integrated for evaluating the given outlier detection model. The performance of different detectors can also be compared through interpretations as COIN provides a unified evaluation basis. COIN can also be applied to existing outlier/anomaly detection methods which already provide explanations for their results. In addition, prior knowledge of attribute characteristics about certain application scenarios can be easily incorporated into the interpretation process, which enables end users to perform model selection according to specific demands. 
The contributions of this work are summarized as follows:
\begin{itemize}[leftmargin=*]
\item We define the interpretation of an outlier as three aspects: abnormal attributes, outlierness score, and the identification of the outlier's local context.
\item We propose a novel model-agnostic framework to interpret outliers, as well as designing a concrete model within the framework to extract interpretation information.
\item Comprehensive evaluations on interpretation quality, as well as case studies, are conducted through experiments on both real-world and synthetic datasets.
\end{itemize}

\begin{table}[t]
\begin{center}
	\begin{tabular}{ c|l }
	\hline \hline
      \multicolumn{1}{c}{\bfseries Notation} & \multicolumn{1}{|c}{\bfseries Definition} \\
    \hline
 	$N$ & the number of data points in the dataset \\
    \hline
 	$M$ & the number of attributes \\
 	\hline
 	$\textbf{x}$ & a data instance, $\textbf{x}\in \mathbb{R}^M$ \\ 
 	\hline
 	$a_m$ & the $m$-th attribute \\
 	\hline
 	$\mathscr{X}$ & all data instances, $\mathscr{X} = \{\textbf{x}_1, \textbf{x}_2, ..., \textbf{x}_N\}$ \\ 
 	\hline
 	$h$ & an outlier detection method \\
 	\hline
 	$\mathscr{O}$ & the collection of detected outliers \\
 	\hline
 	$\textbf{o}_i$ & outlier $i$ identified by the detector \\
 	\hline
 	$\mathscr{O}_i$ & the outlier class corresponding to $\textbf{o}_i$ \\
 	\hline
 	$\mathscr{C}_{i}$ & the context of outlier $\textbf{o}_i$ \\
 	\hline
 	$k$ & the number instances included in $\mathscr{C}_{i}$ \\
 	\hline
 	$s(a_m)$ & suspicious score of attribute $a_m$ \\
 	\hline
 	$d(\textbf{o})$ & outlierness score of $\textbf{o}$ \\
 	\hline \hline
	\end{tabular}
\end{center}
\caption{Symbols and Notations}
\vspace{-20pt}
\label{tb:notation}
\end{table}

\section{Preliminaries}
\textbf{Background}\, Many approaches have been proposed for outlier detection. These approaches can be divided into three categories: density-based, distance-based and model-based. Density-based approaches try to estimate the data distribution, where instances that fall into low-densities regions are returned as outliers~\cite{Breu-etal00lof,Agga-Yu00outlier,Tang-etal02enhancing}. Distance-based methods identify outliers as instances isolated far away from their neighbors~\cite{Knor-etal00distance,Rama-etal00efficient,Bay-etal03mining,Angi-Fass09dolphin,Liu-etal12isolation}. For model-based ones, usually a specific model (e.g., classification, clustering or graphical model) is applied to the data, and outliers are those who do not fit the model well~\cite{Scho-etal01estimating,He-etal03discovering,Tong-CLin11nonnegative}. Other main focuses of relevant research include tackling the challenges of the curse of dimensionality~\cite{Agga-Yu00outlier,Filz-etal08outlier,Krie-etal09clustering}, the massive data volumn~\cite{Rama-etal00efficient,Angi-Fass09dolphin} and heterogeneous information sources~\cite{Pero-Akog16scalable}. However, interpretation of detection results is usually overlooked. Although some recent anomaly detection methods provide explanation with their outcome~\cite{Gao-etal10community,Pero-etal14focused,Liu-etal17accelerated,Liang-etal16robust}, they do not represent all the scenarios. The ignorance of outlier interpretation may lead to several problems. First, for security-related domains, where outlier detection is widely applied, explanations affect whether the results will be accepted by end users. Second, the sparsity of outliers brings uncertainty to evaluation methods. Small disturbance on the detection results may lead to significant variations in evaluation results using traditional metrics such as ROC scores~\cite{davis2006relationship}. Third, it is usually difficult to obtain labels of outliers, so we wonder if it is possible to evaluate the detection performance without ground-truth labels. In this work, we resort to interpretation methods to tackle the challenges above.

\begin{figure}[t]
\centering
\includegraphics[height=1.9in, width=2.63in]{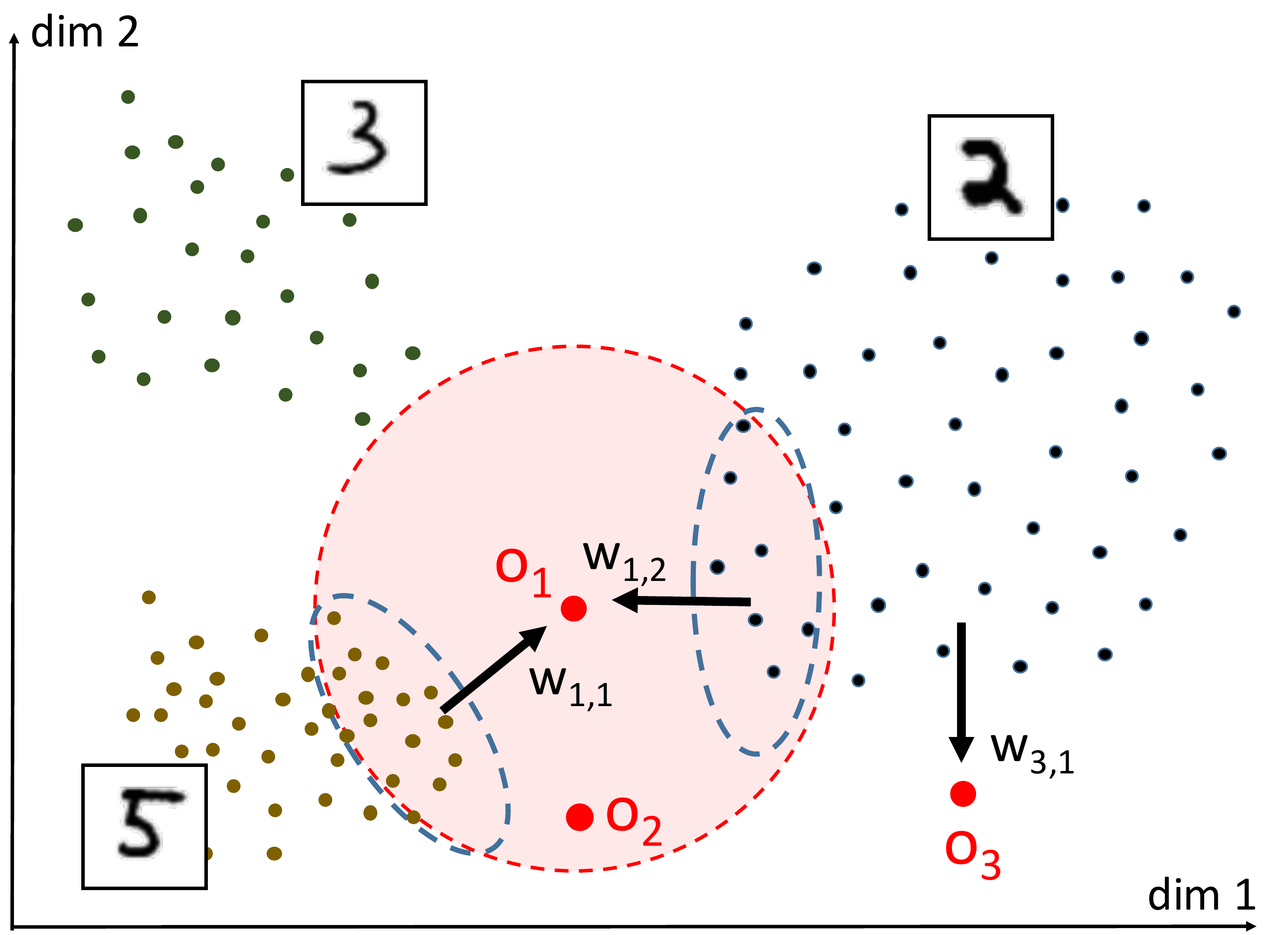}
\caption{A toy example of outlier interpretation by resolving its context in to clusters.}
\vspace{-5pt} 
\label{fig:otlr_clf}
\end{figure}

\noindent \textbf{Notations}\, The notations used in this paper are introduced as below and in Table~\ref{tb:notation}. Let $\mathscr{X}$ denotes the collection of all data. $N$ is the number of data instances in $\mathscr{X}$. Each data instance is denoted as $\textbf{x}\in \mathbb{R}^{M}$, where $M$ is the number of attributes. The $m^{th}$ attribute is denoted as $a_m$. We use $h$ to represent an outlier detector. The collection of outliers identified by a detector is represented as $\mathscr{O}$, in which a single outlier is denoted as $\textbf{o}\in \mathbb{R}^M$. The \textit{context} of an outlier $\textbf{o}$, i.e., $\mathscr{C}_i$, is composed of its $k$-nearest normal instances. Each $\mathscr{C}_i$ could consist of some smaller clusters $\mathscr{C}_{i,1}, \mathscr{C}_{i,2}, ..., \mathscr{C}_{i,L}$. Among the detected outliers, some are far away from the bulks of the dataset, while others are just marginally abnormal. We define the degree of outlierness for an instance $\textbf{x}$ as \textit{outlierness} denoted as $d(\textbf{x})\in \mathbb{R}_{\ge 0}$. The reason for clustering the context is illustrated in Figure~\ref{fig:otlr_clf}. There are three clusters, each of which represents images of a digit. Red points are outliers detected by a certain algorithm. Clusters of digit ``2" and ``5" compose the context of outlier $\textbf{o}_1$. The interpretation of $\textbf{o}_1$, denoted as $\textbf{w}_{1,1}$ and $\textbf{w}_{1,2}$, can be obtained by contrasting it with the two clusters respectively. However, it would difficult to explain the outlierness of $\textbf{o}_1$ if clusters of digit ``2" and ``5" are not differentiated.

\noindent \textbf{Problem Definition}\, Based on the analysis above, here we formally define the outlier interpretation problem as follows. Given a dataset $\mathscr{X}$ and the query outliers $\mathscr{O}$ detected therefrom, the \textit{interpretation} for each outlier $\textbf{o}_i\in \mathscr{O}$ is defined as a composite set: $\mathscr{E}_i = \{\,\mathscr{A}_{i}, d(\textbf{o}_i), \mathscr{C}_{i}=\{\mathscr{C}_{i,l}|l\in [1,L]\} \,\}$. Here $\mathscr{A}_{i}$ include the abnormal attributes of $\textbf{o}_i$ with respect to $\mathscr{C}_{i}$, $d(\textbf{o}_i)$ is the outlierness score of $\textbf{o}_i$, $\mathscr{C}_{i}$ denotes the context of $\textbf{o}_i$ and $\mathscr{C}_{i,l}$ is the $l$-th cluster.


\begin{figure*}[t]
\centering
\includegraphics[height=2.2in, width=6.5in]{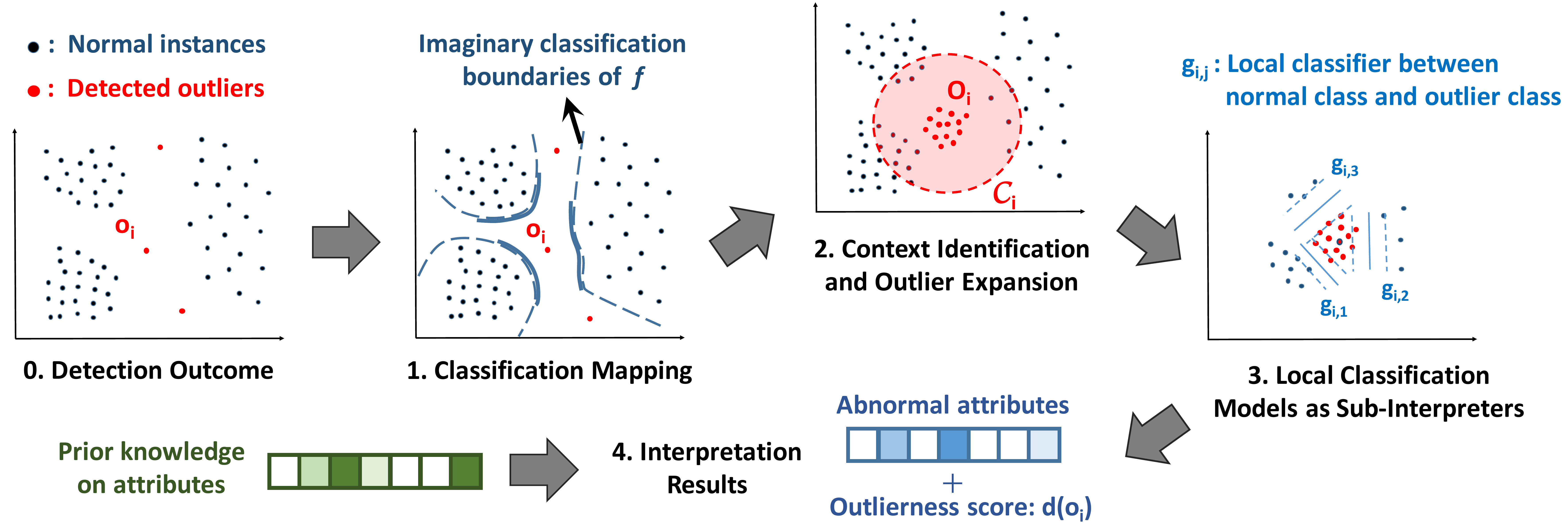}
\caption{Contextual Outlier Interpretation Framework} 
\label{fig:pipeline}
\end{figure*}

\section{Contextual Outlier Interpretation Framework}
The general framework of Contextual Outlier INterpretation (\coin) is illustrated in Figure~\ref{fig:pipeline}. Given a dataset $\mathscr{X}$ and detected outliers $\mathscr{O}$, we first map the interpretation task to a classification problem due to their similar natures. Second, the classification problem over the whole data is partitioned to a series of regional problems focused on the context of each outlier query. Third, a collection of simple and local interpreters $g$ are built around the outlier. At last, the outlying attributes and outlierness score of each outlier can be directly obtained from the parameters of $g$, by combining the application-related prior knowledge. The details of each step are discussed in the following subsections.

\subsection{Explain Outlier Detector with Classifiers}\label{subsec:classfi}
In this module, we will establish the correlation between outlier detection and classification. The close relationship between the two types of problems motivates us to design the interpretation framework from the classification perspective.

Formally, an outlier detection function can be denoted as $h(\textbf{x}|\theta, \mathscr{X})$, where $\textbf{x}\in \mathscr{X}$, $\theta$ and $\mathscr{X}$ represent the function parameters. Here the dataset $\mathscr{X}$ is treated as parameters since data instances affect the degree of normality of each other. The abnormality of an instance is typically represented by either a binary label or a continuous score. In the former case, an instance is categorized as either normal or abnormal, while the latter expresses the degree to which an instance is abnormal. The latter case can be easily transformed to the former if a threshold is set as the separating mark between inlier and outlier classes~\cite{Agga-Yu00outlier, Gao-etal10community, Liu-etal12isolation}. This form of binary detection motivates us to analyze the mechanism of outlier detectors using classification models. Although outlier detection is usually tackled as an unsupervised learning problem, we can assume there exists a latent hyperplane specified by certain function $f(\textbf{x}|\theta'): \mathbb{R}^M \rightarrow \{0, 1\}$ that separates outliers from normal instances. Here $\theta'$ represents the parameters of $f$. This connection between outlier detection and supervised learning has also been implied in some previous work~\cite{Scho-etal01estimating,Abe-etal06outlier}. An intuitive example can be found in Step 1 of Figure \ref{fig:pipeline}. Blue points represent normal instances, and red points indicate the detected outliers. The decision boundaries defined by $f$ are shown using dotted curves. In this setting, the outlier detector is actually trying to mimic the behavior of the decision function.

Given the outliers $\mathscr{O}$ identified by detector $h$, we want to recover the implicit decision function $f$ which leads to the similar detection results as $h$. The problem is thus formulated as below,
\begin{equation}
\operatorname*{arg\,min}_{f} \text{\,\,} \mathcal{L} (h, f; \mathscr{O}, \mathscr{X}-\mathscr{O}) ,
\end{equation}
where $\mathcal{L}$ is the loss function that includes all the factors (e.g., classification error and simplicity of $f$) we would like to consider. $\mathscr{O}$ and $\mathscr{X}-\mathscr{O}$ represent outlier class and inlier class, respectively. However, the final form of $f$ could be very complicated if outliers have diverse abnormal patterns and the whole dataset contains complex cluster structures. Such complexity prevents $f$ from directly providing intuitive explanations for the detector. This is also a common issue in many supervised learning tasks, where the highly complicated prediction function makes the classification model almost a black box. A straightforward solution is to first obtain $f$ and then interpret it~\cite{Baeh-etal10explain,Ribe-etal16whyshould}. This pipeline, however, will introduce new errors in the intermediate steps, and it is more computationally expensive to deal with large datasets. An approach for directly interpreting outlier detectors is needed.

\subsection{Local Interpretation for Individual Outliers}
By utilizing the isolation property of outliers, we can decompose the overall problem of detector interpretation into multiple regional tasks of explaining individual outliers:
\begin{equation}\label{eq:decomp}
\begin{split}
\min_{f}\, \mathcal{L} (h, f; \mathscr{O}, \mathscr{X}-\mathscr{O}) &\Rightarrow \min_{f}\, \sum_i \mathcal{L} (h, f; \textbf{o}_i, \mathscr{C}_{i}) \\
&\Rightarrow \sum_{i\in [1, |\mathscr{O}|]} \min_{g_i}\, \mathcal{L} (h, g_i; \textbf{o}_i, \mathscr{C}_{i}) \\
&\Rightarrow \sum_{i\in [1, |\mathscr{O}|]} \min_{g_i}\, \mathcal{L} (h, g_i; \mathscr{O}_i, \mathscr{C}_{i}).
\end{split}
\end{equation}
In this way, the original problem is transformed to explaining each outlier $\textbf{o}_i$ with respect to its context counterpart $\mathscr{C}_{i}$. Since the number of outliers is usually small, we avoid dealing with the whole dataset which could be large. Here $g_i$ represents the local parts of $f$ exclusively for classifying $\textbf{o}_i$ and $\mathscr{C}_{i}$. In Figure~\ref{fig:pipeline}, for example, $g_i$ is highlighted by the bold boundaries around $\textbf{o}_1$ in Step 1, and $\mathscr{C}_{i}$ consists of the normal instances enclosed in the circle in Step 2. Since there is a data imbalance between the two classes, by applying strategies such as synthetic sampling~\cite{He-Garc09learning}, $\textbf{o}_i$ is expanded to a hypothetical outlier class $\mathscr{O}_i$ with comparable size to $\mathscr{C}_{i}$. As it is common for outlier detectors to measure the outlierness of instances based on their contexts, a proper interpretation method would better take this into consideration.


\subsection{Resolve Context for Outlier Explanations}\label{sec:sep2dist}
Now we focus on interpreting each single outlier $\textbf{o}_i$ by solving $g_i$ in Equation \ref{eq:decomp}. 
Since $g_i$ is the local classifier separating $\mathscr{O}_i$ from $\mathscr{C}_i$, the current task is turned into interpreting the classification boundary of $g_i$. Let $p_{\mathscr{O}_i}(\textbf{x})$ and $p_{\mathscr{C}_i}(\textbf{x})$ denote the probability density function for the outlier class and inlier class, respectively. Since the context $\mathscr{C}_i$ for different $i$ could have various cluster structures as shown in Figure~\ref{fig:otlr_clf}, it is difficult to directly measure the degree of separation between $\mathscr{O}_i$ and $\mathscr{C}_i$ or to discover the attributes that characterize the differences between the two classes. Therefore, we further resolve $\mathcal{L} (h, g_i; \mathscr{O}_i, \mathscr{C}_{i})$ to a set of simpler problems. According to Bayesian decision theory, the classification error equals to
\begin{align*}
    & P^{err}(\mathscr{O}_i, \mathscr{C}_i) \numberthis \label{eq:p2d} \\
=\, & P(\mathscr{O}_i) \int_{\mathscr{C}_i} p(\textbf{x}|\mathscr{O}_i) d\textbf{x} + P(\mathscr{C}_i) \int_{\mathscr{O}_i} p(\textbf{x}|\mathscr{C}_i) d\textbf{x} \\
\approx \, & \big(\sum_{l\in [1,L]} P(\mathscr{O}_i) \int_{\mathscr{C}_{i,l}} p(\textbf{x}|\mathscr{O}_i) d\textbf{x} \,\big) + \big(\sum_{l\in [1,L]} P(\mathscr{C}_{i,l}) \int_{\mathscr{O}_i} p(\textbf{x}|\mathscr{C}_{i,l}) d\textbf{x} \,\big) \\
=\, & \sum_{l\in [1,L]} \big(P(\mathscr{O}_i) \int_{\mathscr{C}_{i,l}} p(\textbf{x}|\mathscr{O}_i) d\textbf{x} + P(\mathscr{C}_{i,l}) \int_{\mathscr{O}_i} p(\textbf{x}|\mathscr{C}_{i,l}) d\textbf{x} \,\big) \\
=\, & \sum_{l\in [1,L]} P^{err}(\mathscr{O}_{i,l}, \mathscr{C}_{i,l}) .
\end{align*}
Suppose we can split the context $\mathscr{C}_i$ into multiple clusters $\{\mathscr{C}_{i,l}|l\in [1,L]\}$ that are sufficiently separated from each other, then cluster $\mathscr{C}_{i,l}$ is the only dominant class near the decision boundary between $\mathscr{O}_i$ and $\mathscr{C}_{i,l}$. Then each term in the summation can be treated as an individual sub-problem of classification without mutual inference. By combining Equation~\ref{eq:decomp} and Equation~\ref{eq:p2d}, our interpretation tasks is finally formulated as:
\begin{equation}\label{eq:decomp2}
\min_{f} \mathcal{L} (h, f; \mathscr{O}, \mathscr{X}-\mathscr{O}) \Rightarrow \min_{g_{i,l}} \sum_i \sum_l \mathcal{L} (h, g_{i,l}; \mathscr{O}_{i,l}, \mathscr{C}_{i,l}).
\end{equation}
By now we are able to classify $\mathscr{O}_{i,l}$ and $\mathscr{C}_{i,l}$ with a simple and explainable model $g_{i,l}$ such as linear models and decision trees, where the outlying attributes $\mathscr{A}_{i,l}$ can be extracted from \textit{model parameters}~\cite{Ribe-etal16whyshould,Lakk-etal16interpretable}. The overall interpretation for $\textbf{o}_i$ can be obtained by integrating the results across all context clusters $\mathscr{C}_{i,l}$, $l\in [1,L]$.

The estimated time complexity for implementing the framework above is $O(|\mathscr{O}| \times L \times T_g)$, where $T_g$ is the average time cost of constructing $g_{i,l}$. Due to the scarcity of outliers, $|\mathscr{O}|$ is expected to be small. Each $g_{i,l}$ involves $\mathscr{O}_{i,l}$ and $\mathscr{C}_{i,l}$. Since $\mathscr{C}_{i,l}$ is only a small subset of data points around an outlier, and $\mathscr{O}_{i,l}$ has comparable size with $\mathscr{C}_{i,l}$, both of their cardinalities should be small, which significantly reduces the time $T_g$. Moreover, the interpretation processes of different outliers are independent of each other, thus can be implemented in parallel to further reduce the time cost. 

\section{Distilling Interpretation from Models}
After introducing the general framework of {\coin}, we have resolved the vague problem of outlier interpretation into a collection of classification tasks around individual outliers. In this section, we will propose concrete solutions for explaining an individual outlier, including discovering its \textit{abnormal attributes} and measuring the \textit{outlierness score}.

\subsection{Context Identification and Clustering}
Given an outlier $\textbf{o}_i$ spotted by a detector $h$, first we need to identify its context $\mathscr{C}_{i}$ in the data space. As introduced before, $\mathscr{C}_{i}$ consists of the nearest neighbors of $\textbf{o}_i$. Here we use Euclidean distance as the point-to-point distance measure. The neighbors are chosen only from normal instances. The instances in $\mathscr{C}_{i}$ are regarded as the representatives for the local background around the outlier. Although $\mathscr{C}_{i}$ contains only a small number of data instances compared to the size of the whole dataset, they constitute the border regions of the inlier class and thus are adequate to discriminate between inlier and outlier classes, as shown in the Step 2 of Figure~\ref{fig:pipeline}.

As local context may indicate some interesting structures (e.g., instances with similar semantics are located close to each other in the attribute space), we further segment $\mathscr{C}_{i}$ into multiple disjoint clusters.
To determine the number of clusters $L$ in $\mathscr{C}_{i}$, we adopt the measure of \textit{prediction strength}~\cite{Tibs-Walt05cluster} which shows good performance even when dealing with high-dimensional data. After choosing the appropriate value for $L$, common clustering algorithms such as K-means or hierarchical clustering are applied to divide $\mathscr{C}_{i}$ into multiple clusters as $\mathscr{C}_{i} = \{\mathscr{C}_{i,1}, \mathscr{C}_{i,2}, \cdots, \mathscr{C}_{i,L}\}$. Clusters of small size, i.e., $|\mathscr{C}_{i,l}| \le 0.03\cdot|\mathscr{C}_{i}|$, are abandoned in subsequent procedures. 

\subsection{Maximal-Margin Linear Explanations}\label{sec:svm}
The concrete type of models chosen for $g_{i,l}$ should have the following properties. First, it is desirable to keep $g\in G$ simple in form. For example, we may expect the number of non-zero weights to be small for linear models, or the rules to be concise in decision trees~\cite{Ribe-etal16whyshould}. Here we let $g\in G$ belong to linear models, i.e., $g(\textbf{x}) = \textbf{w}^T\textbf{x}$. We impose the $l_1$-norm constraint on $\textbf{w}$, where attributes $a_m$ that correspond to large $|w[m]|$ values are regarded as abnormal. Second, since outliers are usually highly isolated from their context, there could be multiple solutions all of which could classify the outliers and inliers almost perfectly, but we want to choose the one that best reflect such isolation property. This motivates us to choose $l_1$ norm support vector machine~\cite{Zhu-etal04l1norm} to build $g$. The local loss $\mathcal{L} (h, g_{i,l}; \mathscr{O}_{i,l}, \mathscr{C}_{i,l})$ to be minimized in Equation (\ref{eq:decomp2}) is thus as below:
\begin{equation}
\begin{split}
  & \sum_{n=1}^{N_{i,l}} (1 - y_n g(\textbf{x}_n) - \xi_n)_+ + c \sum_{n=1}^{N_{i,l}} \xi_n ,\\
\text{s.t.} \quad & \xi_n \ge 0, 
	 		\quad \|\textbf{w}\|_1 \le b
\end{split}
\vspace{-10pt}
\end{equation}
where $N_{i,l} = |\mathscr{O}_{i,l} \cup \mathscr{C}_{i,l}|$, $(.)_+$ is the hinge loss, $\xi_n$ is the slack variable, $b$ and $c$ are the parameters. Here $y_n = 1$ if $\textbf{x}_n\in \mathscr{C}_{i,l}$ and $y_n = -1$ if $\textbf{x}_n\in \mathscr{O}_{i,l}$.

\begin{figure}[t]
\centering
\includegraphics[width=3.3in]{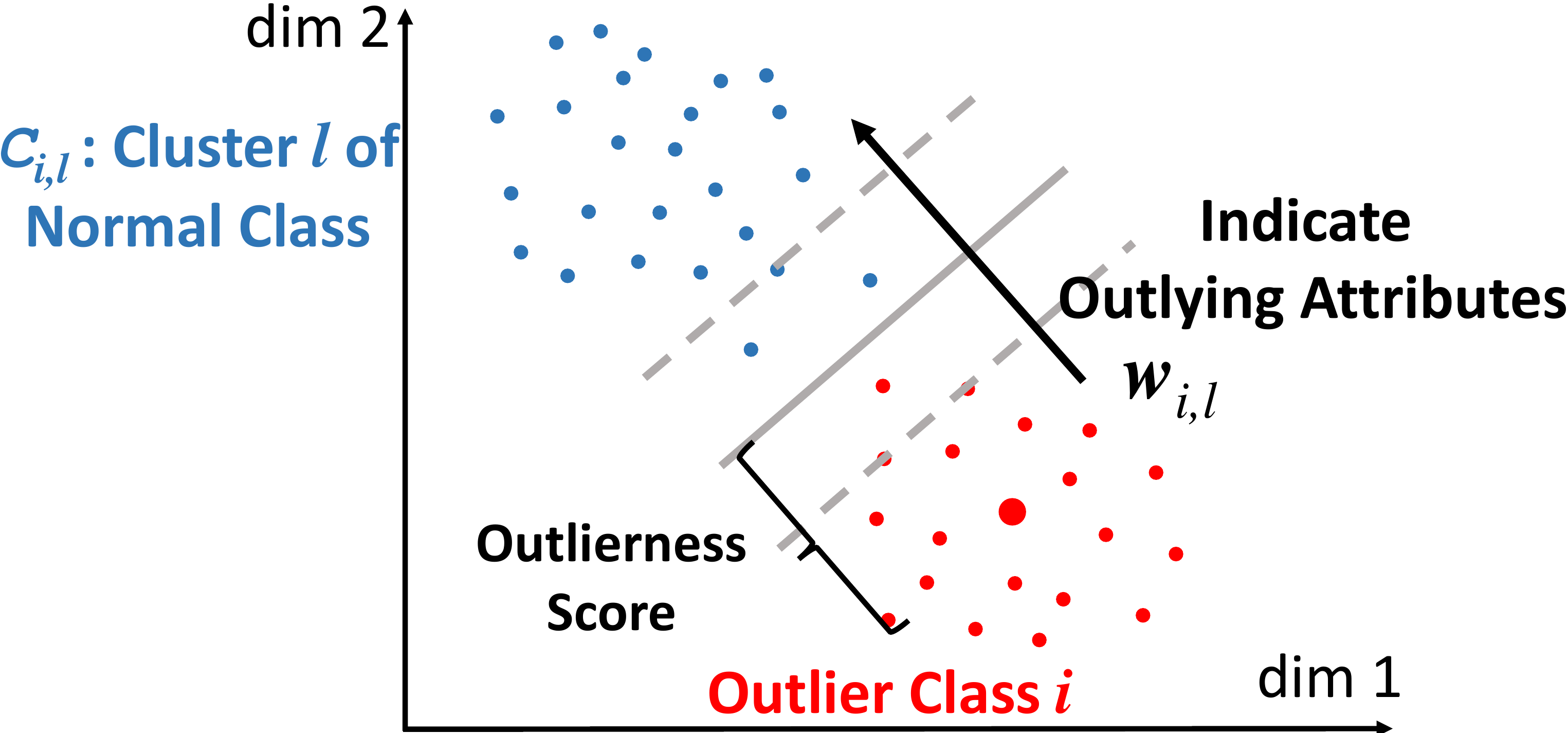}
\caption{Outlier Interpretation from SVM Parameters}
\label{fig:svm}
\end{figure}

From the parameters of the local model $g_{i,l}$, we can find the abnormal attributes and compute the outlierness score with respect to $\mathscr{C}_{i,l}$. Let $\textbf{w}_{i,l}$ denote the weight vector of $g_{i,l}$, the importance of attribute $a_m$ with respect to the context of $\mathscr{C}_{i,l}$ is thus defined as $s_{i,l}(a_m) = |w_{i,l}[m]|/\gamma^m_{i,l}$. Here $\gamma^m_{i,l}$ denotes the resolution of attribute $a_m$ in $\mathscr{C}_{i,l}$, i.e., the average distance along the $m^{th}$ axis between an instance in $\mathscr{C}_{i,l}$ and its closest neighbors.
The overall score of $a_m$ for $\textbf{o}_i$ is
\begin{equation}
s_i(a_m) = (1/|\mathscr{C}_i|) \sum_{l} |\mathscr{C}_{i,l}| s_{i,l}(a_m) ,
\end{equation}
which is the weighted average score for $a_m$ over all $L$ clusters. Attributes $a_m$ with large $s_i(a_m)$ are regarded as the abnormal attributes for $\textbf{o}_i$ (i.e., $a_m\in \mathscr{A}_{i}$). For the outlierness score $d(\textbf{o}_i)$, we define it as:
\begin{equation}
d_l(\textbf{o}_i)=|g_{i,l}(\textbf{o}_i)|/\|\textbf{w}_{i,l}\|_2 .
\end{equation}
This measure is robust to high dimensional data, as $\textbf{w}$ is sparse and $d_l(\textbf{o}_i)$ is calculated in a low dimensional space.
An example is shown in Figure~\ref{fig:svm}, where abnormal attributes are indicated from weight vector $\textbf{w}$ and the outlierness score is computed from the margin of SVM. 
The overall outlierness score for $\textbf{o}_i$ across all context clusters is:
\begin{equation}
d(\textbf{o}_i) = (1/|\mathscr{C}_i|) \sum_l |\mathscr{C}_{i,l}|\, d_l(\textbf{o}_i) / \gamma_{i,l} ,
\end{equation}
which is the weighted summation over different context clusters. Here the normalization term $\gamma_{i,l}$ is the average distance from an instance to its closest neighbor in $\mathscr{C}_{i,l}$. 

\subsection{Incorporate Prior Knowledge into Interpretation}
In real-world applications, the importance of different attributes varies according to different scenarios~\cite{Yang-etal11free, canali2011prophiler, ntoulas2006detecting}. Take Twitter spammer detection as an example. We discuss two attributes of users: the number of followers ($N_{fer}$) and the ratio of tweets posted by API ($R_{API}$). A spammer tends to have small $N_{fer}$ value as they are socially inactive, but large $R_{API}$ in order to conveniently generate malevolent content. However, it is easy for spammers to intentionally increase their $N_{fer}$ by following each other, while manually decreasing $R_{API}$ is more difficult due to the expense human labor. In this sense, $R_{API}$ is more robust and more important than $N_{fer}$ in translating detected outliers as social spammers. To represent the different roles of attributes, we introduce two vectors $\bm{\beta}$ and $\bm{p}$, where $\beta_m\in \mathbb{R}_{\ge 0}$ denotes the relative degree of significance assigned to attribute $a_m$, and $p_m\in \{-1,0,1\}$ denotes the prior knowledge on the expected magnitude of attribute values of outliers. $p_m = -1$ means we expect outliers to have small value for $a_m$ (e.g., $N_{fer}$), $p_m = 1$ means the opposite (e.g., $R_{API}$), while $p_m = 0$ means there is no preference. Therefore, the outlierness score of $\textbf{o}_i$ with respect to $\mathscr{C}_{i,l}$ is refined as:
\begin{equation}
d_l(\textbf{o}_i) = \| \frac{|g_{i,l}(\textbf{o}_i)|}{\gamma_{i,l} \|\textbf{w}_{i,l}\|} \frac{\textbf{w}'_{i,l}}{\|\textbf{w}_{i,l}\|} \circ \bm{\beta}\| ,
\end{equation}
where the operator $\circ$ denotes element-wise multiplication, $w'[m] = \min(0, w[m])$ if $p_m = 1$, and $w'[m] = \max(0, w[m])$ if $p_m = -1$. If we label outliers with $1$ and inliers with $-1$, the sign is reversed. The motivation of introducing $\textbf{w}'$ is that, if interpretation results (e.g., $R_{API}$ is small) does not conform with the expectation expressed by the prior knowledge (e.g., $R_{API}$ is expected to be large to signify spammers), then the outlierness score of the outlier should be deducted. Here $\gamma_{i,l}$ is the average distance from an instance to its closest neighbor in $\mathscr{C}_{i,l}$. It normalizes the outlierness measure with respect to the data density of different clusters. Therefore, the overall outlierness score for $\textbf{o}_i$ across all context clusters is:
\begin{equation}
d(\textbf{o}_i) = \frac{1}{|\mathscr{C}_i|} \sum_l |\mathscr{C}_{i,l}|\, d_l(\textbf{o}_i) ,
\end{equation}
which comprehensively considers the isolation of $\textbf{o}_i$ over different contexts. Now we have obtained all of the three aspects of interpretation $\mathscr{E}_i = \{\,\mathscr{A}_{i}, d(\textbf{o}_i), \mathscr{C}_{i}=\{\mathscr{C}_{i,l}|l\in [1,L]\} \,\}$. If a normal instance is misdetected as an outlier by a detection method, then $\mathscr{E}_i$ is able to identify such mistake, since $d(\textbf{o}_i)$ will be small and $\mathscr{A}_{i}$ is less likely to conform to the prior knowledge.



\section{Experiments}
In this section, we present evaluation results to assess the effectiveness of our framework. We try to answer the following questions: 1) How accurate is the proposed framework in identifying the abnormal attributes of outliers? 2) Can we accurately measure the outlierness score of outliers? 3) How effective is the prior knowledge of attributes in refining outlier detection results? 

\subsection{Datasets}
The real-world datasets used in our experiments include Wisconsin Breast Cancer (WBC) dataset~\cite{ucidata}, MNIST dataset and Twitter spammer dataset~\cite{Yang-etal11free}. The outlier labels are available. WBC dataset records the measurements for breast cancer cases with two classes, i.e. benign and malignant. The former is considered as normal, while we downsampled $25$ malignant cases as the outliers. MNIST dataset includes a collection of $28\times 28$ images of handwritten digits. In our experiments, we only use the training set which contains 42,000 examples. Instead of using raw pixels as attributes, we build a Restricted Boltzmann Machine (RBM) with $150$ latent units to map images to a higher-level attribute space~\cite{Hint-Sala06reducing}. The new low-dimensional attributes are more proper for interpretation than raw pixels. A multi-label logistic classifier is then built to classify different written digits, and the ground truth outliers are selected as the misclassified instances downsampled to $1,000$ of them. The Twitter dataset contains information of normal users and spammers crawled from Twitter. Attributes are classified into two categories according to whether they are robust to the disguise of spammers. Low robustness attributes refer to those which can be easily controlled by spammers to avoid being detected, while high robustness attributes are more trustworthy in discriminating spammers from normal users~\cite{Yang-etal11free}.

We also build two synthetic datasets with ground truth outlying attributes for each outlier, following the procedures in~\cite{Kell-etal12hics}. Both datasets consist of multiple clusters as normal instances generated under multivariate Gaussian distributions. Outliers are created by distorting some samples' attribute values beyond certain clusters, while keeping other attributes within the range of the normal instances. In the first dataset, each outlier is close to only one normal cluster and far away from the others. In the second dataset, an outlier is in the vicinity of several normal clusters simultaneously, while its outlying attributes differ with respect to different neighbors, so that a more refined interpretation approach is desirable.

\begin{table}[t]
\begin{center}
    \begin{tabular}{c|c|c|c|c|c}
      \hline \hline
             & SYN1 & SYN2 & WBC & Twitter & MNIST  \\ \hline
      $N$ & $405$ & $405$ & $458$ & $11$,$000$ & $42$,$000$  \\
      $M$ & $15$ & $15$ & $9$ & $16$ &  $150$ \\
      $|\mathscr{O}|$  & $30$ & $30$ & $25$ & $1$,$000$ & $1$,$000$ \\ \hline
      \hline
    \end{tabular}
\end{center}
        \caption{Details of the datasets in experiments}
\label{table:AUC_real}
\vspace{-25pt}
\end{table}

\subsection{Baseline Methods}
We compare {\coin} with some baseline methods including outlying-aspect mining techniques and classifier interpretation approaches summarized as below:
\begin{itemize}[leftmargin=*]
\vspace{-3pt}
\item CA-lasso (CAL)~\cite{Mice-etal13explaining}: Measure the separability between outlier and inliers as the classification accuracy between the two classes, and then apply feature selection methods (e.g., LASSO) to determine the attribute subspace as explanations.
\item Isolation Path Score with Beam Search (IPS-BS)~\cite{Vinh-etal16discovering}: Apply isolation path score~\cite{Liu-etal12isolation} to measure outlierness. The score is then used to guide the search of subspaces, where Beam Search is applied as the main strategy.
\item LIME~\cite{Ribe-etal16whyshould}: An effective global classification model is first constructed to classify outliers and inliers. Then the outlying attributes for each outlier is identified by locally interpreting the classification model around the outlier. Oversampling is applied to prevent data imbalance. A neural network is used as the global classifier for MNIST data, and SVMs with RBF kernel are used for other datasets.
\end{itemize}

\begin{table*}[t]
\centering
\ra{1.}
\begin{tabular}{@{}rrrrcrrrcrrrcrrr@{}}\toprule
& \multicolumn{3}{c}{\coin} & \phantom{abc}& \multicolumn{3}{c}{CAL} &
\phantom{abc} & \multicolumn{3}{c}{IPS-BS} & \phantom{abc}& \multicolumn{3}{c}{LIME}\\
\cmidrule{2-4} \cmidrule{6-8} \cmidrule{10-12} \cmidrule{14-16}
& Prec & Recall & F1 && Prec & Recall & F1 && Prec & Recall & F1 && Prec & Recall & F1\\ \midrule
SYN1 & {\bf 0.97} & {\bf 0.89} & {\bf 0.93} && 0.89 & 0.81 & 0.84 && 0.87 & 0.44 & 0.58 &&  0.82 & 0.79 & 0.80\\
SYN2 & 0.99 & {\bf 0.90} & {\bf 0.94} && 0.92 & 0.70 & 0.80 && {\bf 1.00} & 0.37 & 0.54 && 0.91 & 0.70 & 0.79\\
WBC & 0.86 & 0.37 & {\bf 0.52} && 0.84 & 0.37 & 0.51 && {\bf 0.90} & 0.15 & 0.26 &&  0.35 & {\bf 0.39} & 0.37\\
Twitter & {\bf 0.91} & 0.33 & 0.48 && 0.75 & 0.34 & 0.47 && 0.72 & 0.29 & 0.41 && 0.60 & {\bf 0.67} & {\bf 0.63}\\
\bottomrule
\end{tabular}
\caption{Faithfulness of Abnormal Attributes Identification} \label{table:exp_attr}
\vspace{-10pt}
\end{table*}

\subsection{Outlying Attributes Evaluation}\label{exp:faith}
The goal of this experiment is to verify that the attributes identified by {\coin} are indeed outlying. Since ground-truth outlying attributes of real-world datasets are not available, we append $M$ noise attributes to all real-world data instances. We simply assume that all of the original attributes are outlying attributes, and noise attributes are not. For each outlier, we apply our approach as well as baseline methods to infer the outlying attributes, and compare the results with the ground truth to evaluate their performances. In our experiments, we choose $8\%$ of nearest neighbors of an outlier $\textbf{o}_i$ as its context $\mathscr{C}_i$. The radius of synthetic sampling for building the outlier class $\mathscr{O}_i$ is set as half of the distance to the inlier class $\mathscr{C}_i$, in order to suppress the overlap between the two classes. The hyperparameters in SVM models are determined through validation, where some samples from $\mathscr{O}_i$ and $\mathscr{C}_i$ are randomly selected as the validation set. The same hyperparameter values are used for all outliers in the same dataset. We report the Precision, Recall and F1 score averaged over all the outliers queries in Table~\ref{table:exp_attr}. Besides finding that {\coin} consistently indicates good performance, some observations can be made as follows:
\begin{itemize}[leftmargin=*]
\item In general, the Recall value of SYN2 is lower than that of SYN1, because the context of each outlier in SYN2  has several clusters, and the true abnormal attributes vary among different clusters. In this case, retrieving all ground-truth attributes is a more challenging task.
\item IPS-BS is relatively more cautious in making decisions. It tends to stop early if the discovered abnormal attributes already make the outlier query well isolated. Therefore, IPS-BS usually achieves high Precision, but only a small portion of true attributes are discovered (low Recall).
\item The Recall scores are low for real-world data since we treat all original attributes to be the ground truth, so low Recall values do not mean bad performances.
\end{itemize}

\subsection{Outlierness Score Evaluation}\label{exp:outscore}
In this experiment, we evaluate if interpretation methods are able to accurately measure the outlierness score of outlier queries. 
An effective outlier detector is less likely to miss instances that are divergent from normal patterns, or consider normal instances to be more suspicious than true outliers. In this regard, the interpretation approach should also be able to accurately measure the degree of deviation of a test outlier from its normal counterpart. In order to simulate ground truth outlierness, for each dataset applied in this experiment, we randomly sample the same number of inliers as the outliers, and use both of them as queries fed into interpreters. The ground truth score is 1 each true outlier, and 0 for inlier samples. For each query instance, interpreters are asked to estimate its outlierness score. After that, we rank the instances in descending order with respect to their scores. A trustworthy interpreter should be able to maintain the relative magnitude of scores among all instances, so true outliers should be assigned with higher scores than inliers.

We report the results in Table \ref{table:exp_outscore} with AUC as the evaluation metric. The proposed method achieves better performance than the baseline methods especially on SYN2 and MNIST. This can be explained by the more complex structures in these datasets, where an outlier may be close to several neighboring clusters.  {\coin} resolves the contextual clusters around each outlier, so it can handle such scenario. This also explains why IPS\_BS is also more effective in complex datasets than the other two baseline methods. The isolation tree used in IPS\_BS can handle complex cluster structures. For SYN1, an outlier is only detached from only one major cluster. For WBC, a malignant instance is usually characterized by those attributes with values significantly larger than normal. The contexts for these two datasets are relatively clear. The performances on these datasets are generally better across all interpreters.

\begin{table}[t!]
\begin{center}
    \begin{tabular}[0.1\textwidth]{c|c|c|c|c|c}
      \hline \hline
      AUC  & SYN1 & SYN2 & WBC & Twitter & MNIST \\ \hline
      \coin & $0.78$ & $0.93$ & $0.96$ & $0.85$ & $0.87$ \\
      CAL & $0.71$ & $0.63$ & $0.94$ & $0.81$ & $0.76$ \\
      IPS\_BS  & $0.69$ & $0.91$ & $0.90$ & $0.79$ & $0.82$ \\
      LIME  & $0.74$ & $0.62$ & $0.94$ & $0.83$ & $0.78$ \\ \hline
      \hline
    \end{tabular}
\end{center}
        \caption{Outlierness score ranking performance} \label{table:exp_outscore}
        \vspace{-2pt}
\end{table}


\subsection{Interactions between Outlying Attributes and Outlierness}\label{exp:inter}
In real-world scenarios, outlier detection may serve for some practical purposes, such as spammer detection, fraud detection and health monitoring. From the outlying attributes revealed by interpretation models, base on human knowledge, we can judge if their roles or semantics are in accordance with the nature of the problem. For those outliers whose abnormal attributes are loosely related to the problem, we want to weaken their significance or even discard them. In this experiment, we discuss how to refine the outlier detection results in terms of increasing the relevancy between spotted outliers and applications, by incorporating prior knowledge of the practical meaning of attributes.

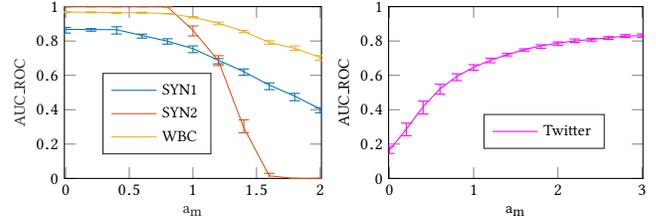
\begin{figure}[t]
\captionsetup[subfigure]{justification=centering}
\hspace{-12pt}
\begin{subfigure}[b]{0.22\textwidth}
  \setlength\figureheight{0.9in}
  \setlength\figurewidth{1.4in}
  \centering  \scriptsize
%
%
\definecolor{mycolor1}{rgb}{0.00000,0.44700,0.74100}%
\definecolor{mycolor2}{rgb}{0.85000,0.32500,0.09800}%
\definecolor{mycolor3}{rgb}{0.92900,0.69400,0.12500}%
\begin{tikzpicture}

\begin{axis}[%
width=0.961\figurewidth,
height=\figureheight,
at={(0\figurewidth,0\figureheight)},
scale only axis,
xmin=-0.01,
xmax=2.01,
xlabel style={font=\color{white!15!black}},
xlabel={$\text{a}_\text{m}$},
ymin=0,
ymax=1,
ylabel style={font=\color{white!15!black}},
ylabel={AUC\_ROC},
axis background/.style={fill=white},
legend style={at={(0.148,0.14)}, anchor=south west, legend cell align=left, align=left, draw=white!15!black}
]
\addplot [color=mycolor1]
 plot [error bars/.cd, y dir = both, y explicit]
 table[row sep=crcr, y error plus index=2, y error minus index=3]{%
0	0.867888888888889	-0.0215555555555557	-0.0106666666666666\\
0.2	0.866444444444445	-0.00911111111111107	-0.00588888888888905\\
0.4	0.865888888888889	-0.0252222222222221	-0.017\\
0.6	0.832222222222222	-0.0155555555555554	-0.00777777777777777\\
0.8	0.798777777777778	-0.01677777777777777	-0.0143333333333333\\
1	0.755555555555556	-0.0233333333333333	-0.0161111111111112\\
1.2	0.688111111111111	-0.0207777777777778	-0.00977777777777777\\
1.4	0.621333333333333	-0.020888888888889	-0.0157777777777778\\
1.6	0.543666666666667	-0.0268888888888889	-0.0125555555555557\\
1.8	0.479222222222222	-0.0268888888888889	-0.0158888888888889\\
2	0.404222222222222	-0.0218888888888888	-0.00755555555555559\\
};
\addlegendentry{SYN1}

\addplot [color=mycolor2]
 plot [error bars/.cd, y dir = both, y explicit]
 table[row sep=crcr, y error plus index=2, y error minus index=3]{%
0	1	0	0\\
0.2	1	0	0\\
0.4	1	0	0\\
0.6	1	0	0\\
0.8	0.999047619047619	-0.000952380952380927	-0.00238095238095248\\
1	0.862698412698413	-0.0328571428571428	-0.0249206349206351\\
1.2	0.684761904761905	-0.0280158730158731	-0.0297619047619047\\
1.4	0.294761904761905	-0.0291269841269841	-0.0464285714285715\\
1.6	0.0150793650793651	-0.0226984126984127	-0.0150793650793651\\
1.8	0	0	0\\
2	0	0	0\\
};
\addlegendentry{SYN2}

\addplot [color=mycolor3]
 plot [error bars/.cd, y dir = both, y explicit]
 table[row sep=crcr, y error plus index=2, y error minus index=3]{%
0	0.96816	-0.00303999999999993	-0.00176000000000021\\
0.2	0.96768	-0.00352000000000008	-0.00207999999999986\\
0.4	0.96448	-0.00511999999999979	-0.00288000000000022\\
0.6	0.9648	-0.004	-0.00319999999999998\\
0.8	0.95904	-0.00095999999999985	-0.000640000000000085\\
1	0.93872	-0.00448000000000004	-0.00351999999999997\\
1.2	0.90368	-0.00912000000000013	-0.00607999999999997\\
1.4	0.85648	-0.00751999999999997	-0.00528000000000006\\
1.6	0.79136	-0.01024	-0.0105600000000001\\
1.8	0.7576	-0.0120000000000001	-0.0135999999999999\\
2	0.70464	-0.01696	-0.0102400000000001\\
};
\addlegendentry{WBC}

\end{axis}
\end{tikzpicture}%
  \caption{Data with noise attributes}
  \label{fig:attr_score1}
\end{subfigure}%
\hspace{9pt}
\begin{subfigure}[b]{0.22\textwidth}
  \setlength\figureheight{0.9in}
  \setlength\figurewidth{1.4in}
  \centering  \scriptsize
%
%
\definecolor{mycolor1}{rgb}{1.00000,0.00000,1.00000}%
\begin{tikzpicture}

\begin{axis}[%
width=0.951\figurewidth,
height=\figureheight,
at={(0\figurewidth,0\figureheight)},
scale only axis,
xmin=-0.01,
xmax=3.01,
xlabel={$\text{a}_\text{m}$},
ymin=0,
ymax=1,
ylabel={AUC\_ROC},
axis background/.style={fill=white},
legend style={at={(0.373,0.172)},anchor=south west,legend cell align=left,align=left,draw=white!15!black}
]
\addplot [color=mycolor1,solid]
 plot [error bars/.cd, y dir = both, y explicit]
 table[row sep=crcr, y error plus index=2, y error minus index=3]{%
0	0.1651	-0.0193	-0.0219\\
0.2	0.2866	-0.0382	-0.0354\\
0.4	0.4181	-0.0431	-0.0317\\
0.6	0.5213	-0.0303	-0.0269\\
0.8	0.594	-0.0248	-0.0152\\
1	0.6474	-0.0174	-0.0145999999999999\\
1.2	0.6878	-0.0146000000000001	-0.0122\\
1.4	0.7208	-0.00760000000000005	-0.00719999999999998\\
1.6	0.7499	-0.00849999999999995	-0.00470000000000004\\
1.8	0.772	-0.014	-0.00600000000000001\\
2	0.7877	-0.0123	-0.00690000000000002\\
2.2	0.8008	-0.00839999999999996	-0.0112000000000001\\
2.4	0.809	-0.00939999999999996	-0.00700000000000001\\
2.6	0.8189	-0.0091	-0.00649999999999995\\
2.8	0.8286	-0.00780000000000003	-0.00860000000000005\\
3	0.8312	-0.0092000000000001	-0.0099999999999999\\
};
\addlegendentry{Twitter};

\end{axis}
\end{tikzpicture}%
  \caption{Twitter spammer data}
  \label{fig:attr_score2}
\end{subfigure}
\vspace{-5pt}
\caption{The influence of the prior knowledge on outlierness score. Results averaged over 20 runs, bars depict 25-75$\%$.}
\label{fig:attr_score}
\vspace{-0pt}
\end{figure}

The experiment is separated into two parts. In the first part, we assume that all the original attributes are equally relevant to the problem of interest, while some \textit{simulated attributes} are appended to each instance. Different from the noise attributes in Section~\ref{exp:faith} that are of small magnitude, the attributes here may turn inliers to ``outliers". Similar to the previous experiment, we randomly sample the same number of inliers as test instances in addition to the true outliers. Here we set the number of simulated attributes to be the same as original ones, so each instance is augmented as $\textbf{x}\in \mathbb{R}^{2M}$. We run {\coin} on different significance vectors $\bm{\beta}$ and set all entries in $\textbf{p}$ to be zero. The weights corresponding to original attributes are fixed to $1$ ($\beta_m = 1, m\in [1, M]$), and we only vary the weights of simulated attributes ($\beta_m = \beta, m\in [M+1, 2M]$). Similar to Section \ref{exp:outscore}, we obtain the outlierness score for all queries and rank them in descending order according to the score magnitude. True outliers are expected to have higher ranks than inliers. The performance of outlierness ranking is reported in Figure \ref{fig:attr_score1}. The plot indicates that as we increase the weights of noise attributes, the performance of the interpreter degrades for all datasets, because it is more difficult for the interpreter to distinguish between real outliers and noisy instances. From the opposite perspective, assigning large weights to important attributes will filter out misclassified outliers.

The second part of the experiment uses Twitter dataset which consists of the information of a number of normal users and spammers. The features extracted from user profiles, posts and graph structures are used as attributes. According to~\cite{Yang-etal11free}, the robustness level varies for different attributes. Some attributes, such as the number of followers, hashtage ratio and reply ratio, can be easily controlled by spammers to make themselves look normal, so that they are of low robustness. Other attributes such as account age, API ratio and URL ratio are beyond their easy control due to the huge potential expense or human labor, so they have high robustness. In this experiment, we fix the weight of low-robustness attributes to 1, and vary the weight $\beta_m$ of high-robustness attributes. The entries of $\textbf{p}$ are decided according to~\cite{Yang-etal11free}. The remaining procedures are the same as the first part of experiment discussed above. The number of normal instance queries is the half of the real outliers. The result of outlierness ranking is reported in Figure~\ref{fig:attr_score2}. The rising curve shows that as more emphasis is put on high-robust attributes, we are able to refine the performance of identifying spammers. The experiment result indicates that by resorting to the interpretation of detected outliers, we can gain more insights on their characteristics, and more accurately select those that are in accordance with the purpose of the application.

\begin{figure}[t]
\centering
\includegraphics[height=1.96in, width=3.3in]{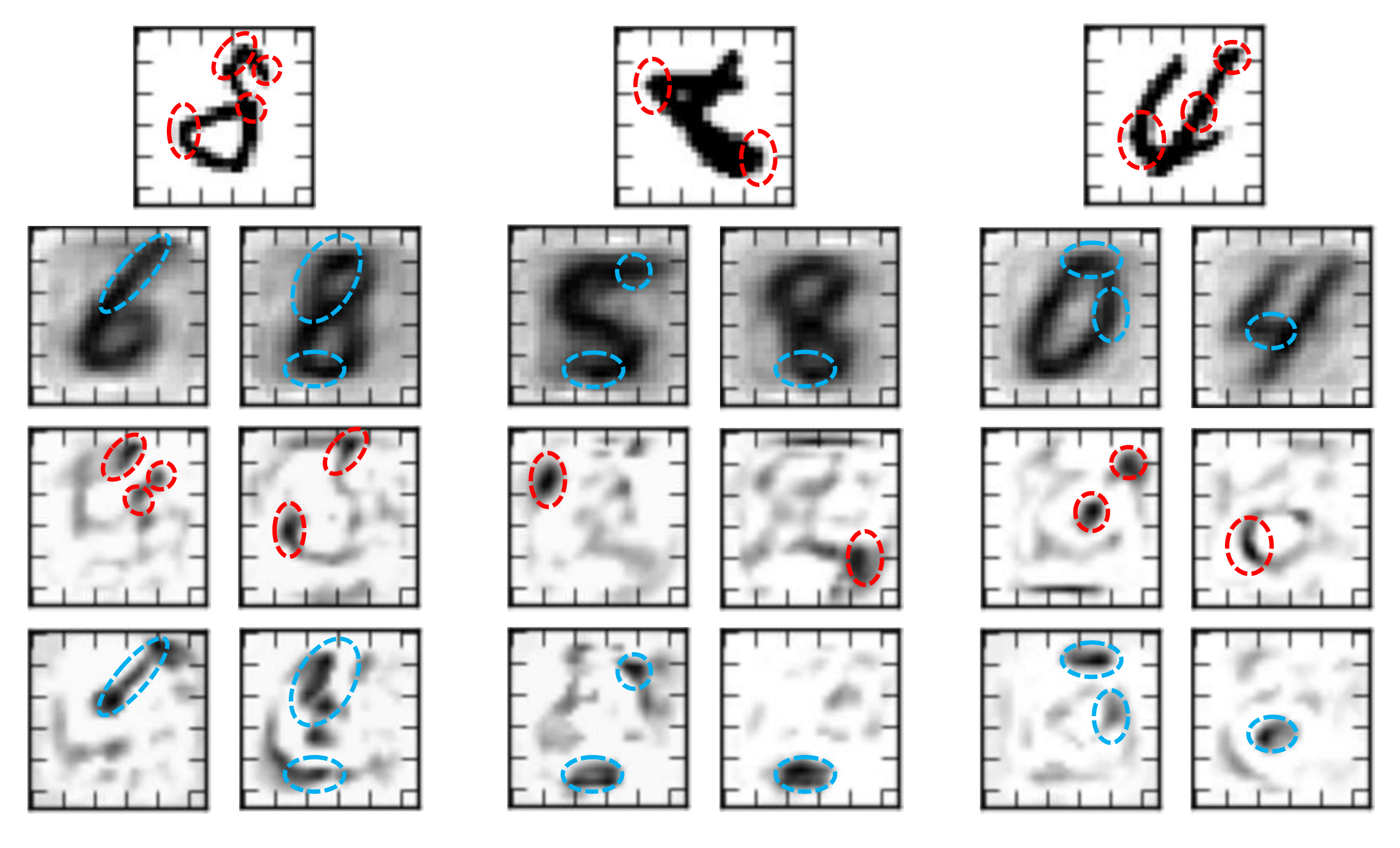}
\vspace{-5pt}
\caption[]
        {\small Examples of outlier interpretation using MNIST dataset. The first row lists the query outlier images. The second row includes the average images over all instances in certain context clusters. Red and blue circles in the third and fourth row highlight the strokes explaining why the query images in the first row are recognized as outliers.}
\label{fig:mnist}
\end{figure}

\subsection{Case Studies}
At last, we conduct some case studies to intuitively present the outcome of different components in {\coin}. MNIST dataset is used here as images are easier to understand perceptually. The attributes fed into the interpreter are hidden features extracted by the RBM. The latent features learned from RBM can be seen as simple primitives that compose more complicated visual patterns. It is more suitable for interpretation than using raw pixels as it is in accordance with the cognitive habits of people, that we tend to use richer representations for explanation and action~\cite{Lake-etal15humanlevel}.

The case study results are shown in Figure \ref{fig:mnist}. There are three query outlier images. The query outlier images are in the first row. We choose two neighboring clusters for each query, and obtain the average image of each cluster, as shown in the second row. Clear handwritten digits can be seen from average images, so that the clusters are internally coherent. The third and fourth rows indicate the characteristic attributes of the query image and average-neighbor image, respectively. The black strokes in the images of the third row represent positive outlying attributes, i.e., the query image is regarded as an outlier instance because it \textit{possesses} these attributes. The strokes in fourth-row images are negative outlying attributes, as the query outlier digit does not include them. These negative attributes are, however, commonly seen in the neighbor images of certain cluster. The positive and negative attributes together explain why the outlier image is different from its nearby images in the dataset.


\section{Related Work}
Many outlier detection approaches have been developed over the past decades. These approaches can be divided into three categories: density-based, distance-based and model-based approaches. Some notable density-based detection methods include~\cite{Breu-etal00lof,Agga-Yu00outlier,Tang-etal02enhancing,Song-etal07conditional,Gao-etal10community}. Representative distance-based approaches include~\cite{Knor-etal00distance,Rama-etal00efficient,Bay-etal03mining,Angi-Fass09dolphin,Liu-etal12isolation}. For model-based approaches, some well-known examples are~\cite{Scho-etal01estimating,He-etal03discovering,Tong-CLin11nonnegative}. Varios approaches have been proposed to tackle the challenges including the curse of dimensionality~\cite{Agga-Yu00outlier,Filz-etal08outlier,Krie-etal09clustering}, the massive data volumn~\cite{Rama-etal00efficient,Angi-Fass09dolphin}, and heterogenous information sources~\cite{Gao-etal10community,Pero-Akog16scalable}. Ensemble learning, which is widely used in supervised learning settings, can also be applied for outlier detection with non-trivial improvements in performance~\cite{Zime-etal13subsampling,Liang-etal16robust}. ~\cite{Laza-Kuma05feature} combines results from multiple outlier detectors, each of which apply only a subset of features. In contrast, each individual detector can subsamples data instances to form a ensemble of detectors~\cite{Zime-etal13subsampling}. Some recent work starts to realize the importance about the explanations of detection results. In heterogeneous network anomaly detection, ~\cite{Gao-etal10community, Pero-etal14focused, Liu-etal17accelerated, Liang-etal16robust} utilize attributes of nodes as auxiliary information for explaining the abnormality of resultant anomaly nodes. The motivation of this work is different from them, as we try to infer the reasons that why the given outliers are regarded as outlying, instead of developing new detection methods.

Besides algorithm development, researcher are also trying to provide explanations along with the approaches and their outcomes. The approach introduced in~\cite{Knor-Ng99findingintensional} can also find the subspace in which the features of outliers are exceptional. Ert\"{o}z \textit{et al.} designed a framework for detecting network intrusion with explainations, which only works on categorical attributes~\cite{Erto-etal04minds}. The Bayesian program learning framework has been proposed for learning visual concepts that generalizes in a way similar to human, especially with just one or a few data examples~\cite{Lake-etal15humanlevel}. Interpretations for anomalies detection can be naturally achieved within the scenario of attributed networks~\cite{Gao-etal10community, Liu-etal17accelerated, Pero-etal14focused}. These techniques cannot be directly applied to solve our problem, because: (1) Heterogenous information may not be available; (2) In many cases, features are not designed for achieving specific tasks; (3) The definition of anomalies varies in the work above, so a more general interpretation approach is still needed. Moreover, given the black-box characteristics of major mathematical models, the community is exploring ways to interprete the mechanisms that support the model, as well as the rules according to which the predictions are made. Ribeiro \textit{et al.} developed a model-agnostic framework that infers explanations by approximating local input-output behavior of the original supervised learning model~\cite{Ribe-etal16whyshould}. Lakkaraju \textit{et al.} formalizes decision set learning which can generate short, succinct and non-overlapping rules for classification tasks~\cite{Lakk-etal16interpretable}. Micenkov\'{a} \textit{et al.} proposed to use classification models and feature selection methods to provide interpretations to the outliers in the subspace~\cite{Mice-etal13explaining}. Vinh \textit{et al.} utilize the isolation property of outliers and apply isolation forest for outlying aspects discovery~\cite{Vinh-etal16discovering}.

\section{Conclusion and Future Work}
In this paper, we propose the Contextual Outlier INterpretation (\coin) framework. The framework is model-agnostic and can be applied to a wide range of detection methods. The goal of interpretation is achieved by solving a series of classification tasks. Each outlier query is explained within its local context. The abnormal attributes and outlierness score of an outlier can be obtained by a collection of simple but interpretable classifiers built in its resolved context. We also propose a new measure of outlierness score whose relationship with abnormal attributes can be explicitly formulated. Prior knowledge on the roles of attribute in different scenarios can also be easily incorporated into the interpretation process. The explanatory information of multiple queries can be aggregated for evaluating detection models. Comprehensive evaluation on interpretation performance and model selection accuracy are provided through a series of experiments with both real world and simulated datasets. Case studies are also conducted for illustrating the outcome of each component of the framework.

There are a number of directions for future work that can be further explored. Hierarchical clustering strategies can be designed to more accurately resolve of the context of an outlier query for better interpretation. The framework can be extended to handle heterogeneous data sources. Moreover, strategies for dealing with outlier groups can be designed, so that interpretation approaches can be applied to a wider range of objects.


%
\bibliographystyle{ACM-Reference-Format}
\bibliography{ninghao_WWW18_ref}
\end{document}